\def\year{2019}\relax
\documentclass[letterpaper]{article} 
\usepackage{aaai19}  
\usepackage{times}  
\usepackage{helvet} 
\usepackage{courier}  
\usepackage[hyphens]{url}  
\usepackage{graphicx} 
\usepackage{times}
\usepackage{epsfig}
\usepackage{amsmath}
\usepackage{color}
\usepackage{amssymb}
\usepackage{color}
\usepackage{multirow}
\usepackage{xparse}

\usepackage[draft]{hyperref}       
\usepackage{url}            
\usepackage{booktabs}       
\usepackage{amsfonts}       
\usepackage{nicefrac}       
\usepackage{microtype} 
\usepackage{epstopdf}
\usepackage{rotating,booktabs,multirow}
\usepackage{amsmath,amssymb} 
\usepackage{color}
\usepackage{subcaption}
\usepackage{wrapfig,lipsum}

\usepackage{algorithm}
\usepackage{algorithmic}
\usepackage[algo2e]{algorithm2e}
\usepackage{mathtools}
\usepackage{caption}
\newcommand{\etal}{\textit{et al. }}

\makeatletter
\newcommand*{\rom}[1]{\expandafter\@slowromancap\romannumeral #1@}
\makeatother
\def\UrlFont{\rm}  
\title{Boosting Network Weight Separability via Feed-Backward Reconstruction}
\author{Jongmin Yu\textsuperscript{\rm 1,*} and Hyeontaek Oh\textsuperscript{\rm 1}\thanks{denotes a corresponding author: Jongmin Yu (e-mail: andrew.yu@kaist.ac.kr)}\\ 
\textsuperscript{\rm 1}Institute for IT Convergence, Korea Advanced Institute of Science and Technology (KAIST),\\
291 Daehak-ro, Yuseong-gu, Daejeon, Republic of Korea (34141)\\
\{andrew.yu\textsuperscript{\rm 1,*}, hyeontaek\textsuperscript{\rm 1}\}@kaist.ac.kr\\}
\begin{document}

\maketitle

\begin{abstract}
This paper proposes a new evaluation metric and boosting method for weight separability in neural network design. In contrast to general visual recognition methods designed to encourage both intra-class compactness and inter-class separability of latent features, we focus on estimating linear independence of column vectors in weight matrix and improving the separability of weight vectors. To this end, we propose an evaluation metric for weight separability based on semi-orthogonality of a matrix and Frobenius distance, and the feed-backward reconstruction loss which explicitly encourages weight separability between the column vectors in the weight matrix. The experimental results on image classification and face recognition demonstrate that the weight separability boosting via minimization of feed-backward reconstruction loss can improve the visual recognition performance, hence universally boosting the performance on various visual recognition tasks.   
\end{abstract}

Representation learning based on deep learning methods has been achieved remarkable performances in various visual recognition studies such as image classification \cite{lecun1998gradient,krizhevsky2012imagenet,he2016deep}, object recognition \cite{eitel2015multimodal,socher2012convolutional}, face recognition \cite{schroff2015facenet,sun2014deep,sun2015deepid3,liu2017sphereface}, and person re-identification \cite{li2014deepreid,ding2015deep}. The key of these successes is the effective feature extraction via the non-linear and cascaded kernel structure of deep neural networks. However, in addition to extracting feature using locally connected and shared weight structure of a convolutional neural network, the neural networks' decision metrics based on Euclidean geometry have been demonstrating that embedded features on inner product space are sufficient to achieve superior recognition accuracies to the conventional discriminative approaches \cite{dalal2005histograms,lowe1999object,zhang2006svm} based on hand-crafted features in various recognition tasks. 

In recent years, not only studies to improve the representation learning capabilities of convolutional neural networks based on modifying structures of networks \cite{he2016deep,huang2017densely}, but also the discriminative embedding methods for latent features into Euclidean space have been actively studied  \cite{liu2016large,liu2017sphereface,wen2016discriminative}. Feature learning constrained on $l_{2}$-norm space \cite{taigman2014deepface} is proposed to improve the discriminative power of learned features by regularizing the vector scale of each data point. Angular cost function \cite{deng2018arcface} is presented. Angular cost functions, Large-margin softmax function \cite{liu2016large}, and Sphereface \cite{liu2017sphereface} are proposed to improve the discriminative properties of learned features based on the understanding of the principle of cosine similarity. \cite{wen2016discriminative} presents the 'center loss' based on clustering methodology, and shows that even though the function is non-differential, it can improve the discriminative power of learned features during network training. Intuitively, these approaches are typically concentrated on the embedding latent features into some constrained space using restriction methodologies for the features by reinforcing of intra-class compactness and inter-class separability \cite{liu2016large}. Even though these approaches have achieved remarkable performance in diverse visual recognition tasks, improving separability of learned weight kernels is one of the challenging issues. In recognition tasks by computing vector similarities between weight and latent features, inner product correlation between weight vectors can significantly affect the performance of the recognition models. 

\begin{figure}
	\vspace{-0.2cm}
	\centering
	\includegraphics[width=\columnwidth]{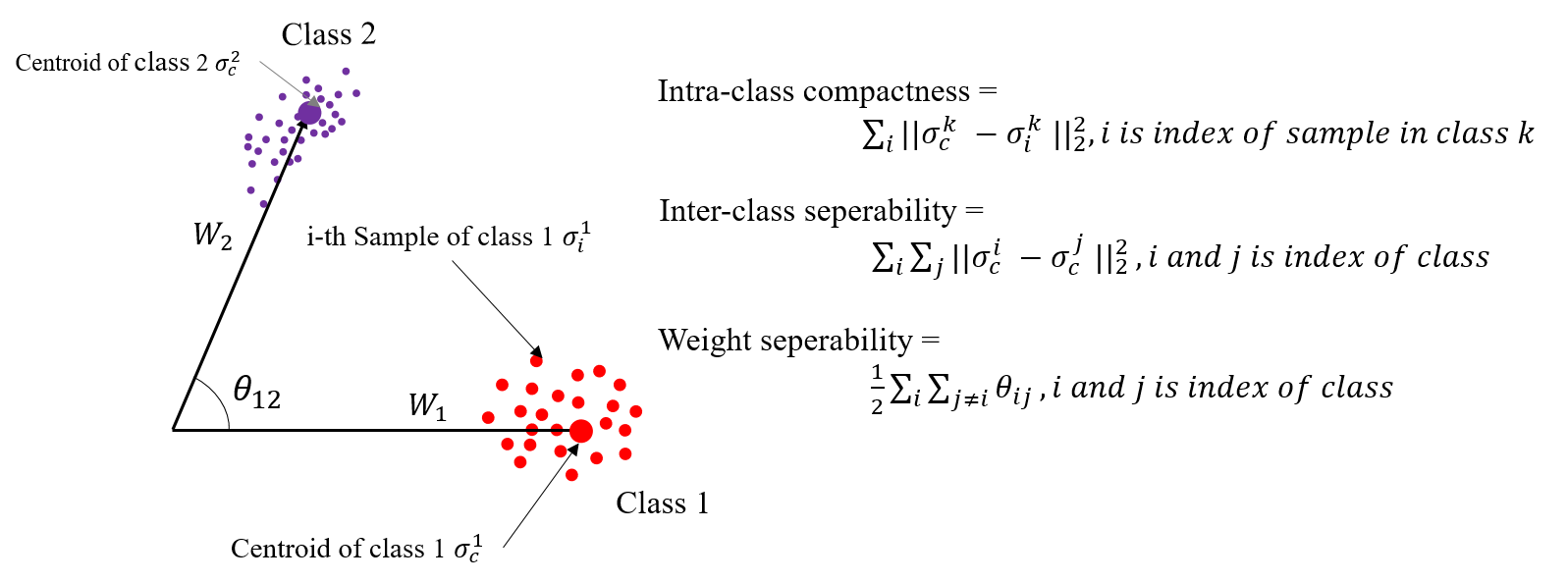}
	\caption{The intuitions of intra-class compactness, inter-class separability, and weight separability. $\sigma^{c}_{j}$ and $\sigma^{k}_{j}$ are the centroid and the $k^{th}$ sample of the $j^{th}$ class. $\theta_{ij}$ is the angle between the $i^{th}$ and $j^{th}$ weight vectors.}
	\label{fig2}
	\vspace{-0.2cm}
\end{figure}

In this paper, we formulate the evaluation metric for weight separability and propose a method to boost the separability of a network weight in a last fully connected layer. Figure \ref{fig2} shows the intuitive concepts of weight separability, inter-class separability, and intra-class compactness. Although one-hot encoded label vectors already induce the weight vectors of last fully connected layer to be orthogonal in general approaches, there is a possibility for further improvement of discriminative power of learned features by revising loss functions or structural details \cite{wen2016discriminative,liu2017sphereface,liu2016large}. We focus on the semi-orthogonalization of a weight matrix, which is a process to find a set of orthogonal vectors that can span a specific subspace. The set of orthogonal vectors takes linear independence between elements. The orthogonalization of weight in neural network is considered as a regularization method to reduce the correlation between detected features by networks \cite{rodriguez2016regularizing}. Our main hypothesis is that the separability between vectors of a weight matrix is related to the recognition performances and it can be evaluated by the linear independence of the weight matrix. The purpose of this paper, therefore, is to prove the hypothesis and apply this intuition to improving representation learning capability of deep neural networks for various visual recognition tasks.

Our key contributions are as follows. First, we define and demonstrate a quantitative evaluation metric for weight separability, which can be used for high-dimensional features without any dimension reduction method and visualization task. Second, we propose a straightforward method to boost the separability of the weight vectors explicitly during network learning. The experimental results show that the proposed method can improve the performance of image classification and face recognition tasks.

\section{Linearity and Separability}
\label{submission}
In commonly used deep learning structures for visual recognition tasks, a fully connected network is used to assign the label by calculating the confidence based on vectorial or probabilistic approaches. The column vectors of weight matrix in last fully connected neural network are used to decide recognition classes of inputs based on the vector similarity based on the inner product: $w_{i}\cdot \alpha=\parallel{w_{i}}\parallel\parallel{\alpha}\parallel\cos\theta_{i}$, where $w_{i}$ is the $i^{th}$ column vector of weight matrix $W=[w_{1}, w_{2}, w_{3}, ..., w_{n}]\in R^{m\times n}$ where $m$ and $n$ is the row and column dimensionalities of weight matrix, and $\alpha$ and $\theta_{i}$ are a latent feature vector and the angle between $w_{i}$ and $a$ respectively. In fully connected networks positioned at the last layer, the figures $m$ and $n$ indicate that the dimensionality of input feature and the number of classes. In recognition task using fully connected layer, the class of a latent feature is assigned as the index of column vector which takes the largest value calculated by the inner product defined as follows:
\begin{equation}
\begin{aligned}
ID = \text{argmax}_{i}(\alpha\cdot{}w_{i}+b), 
\end{aligned}
\end{equation}
where $i$ is the index of column vectors in a weight matrix, $alpha$ is a latent feature. $w_{i}$ and $b$ are $i^{th}$ column vector in the weight matrix and a bias term respectively. The left side terms of above fomular can be changed like a $\text{argmax}_{i}(f(\alpha\cdot{}w_{i}+b))$, where $f$ is an activation function in a network. We omit the bias term and use the augmented vector form to simplify the experiment process. In this paper, we argue that linear independence of the column vectors in a weight matrix has a relation to the separability of weight vectors which can influence performance of various recognition tasks based on vector similarities. To justify our argumentation, we conduct a simple experiment using MNIST dataset \cite{lecun1998gradient}. In these experiments, we used samples of classes: 0,1, and 5 only. We compare two neural networks which have the same structure but trained in different ways. We have employed LeNet \cite{lecun1998gradient} structure in our experiment. One network is trained by forcing with linearly dependent column vectors, and the other is composed of linearly independent column vectors in a final layer. We initially assign random real numbers between $-1$ to $1$, and conduct QR decomposition to take the weight matrix composed of linearly independent column vectors. The formula for the above process is represented as follow:
\begin{equation}
\begin{aligned}
W = \hat{W}R, \hat{W}\hat{W}^{T} = \hat{W}^{T}\hat{W} = I,
\end{aligned}
\end{equation}
where $W \in R^{m\times n}$ is randomly initialized weight matrix, and $\hat{W}\in R^{m\times n}$ is an orthogonal matrix composed of linearly independent column vectors. $R\in R^{n\times n}$ is an upper triangular matrix. We employed a square matrix ($W \in R^{10\times 10}$) in this experiments even though QR decomposition is applied to $m\times{}n$ matrix, with $m\ge{}n$. To maintain the linear independence to the weight vectors during learning, the parameter in the final weight matrix is not updated during training each model.

\begin{figure}
	\vspace{-0.2cm}
	\centering
	\includegraphics[width=\columnwidth]{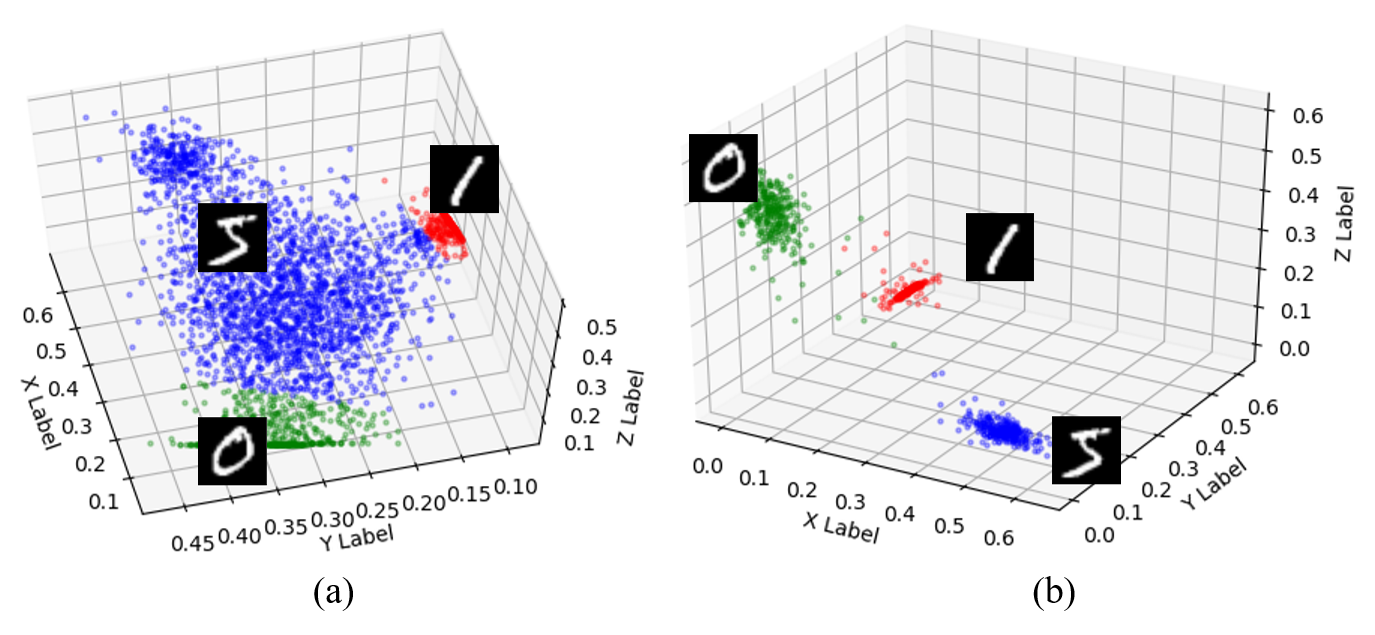}
	\caption{The comparison of the distributions of latent features between the normally trained network (a) and the network (b) which is forced to have the linear independence in their weight matrix. The green, red, and blue points are latent features extracted from input data of the 0,1,5 classes respectively.}
	\label{fig3}
	\vspace{-0.2cm}
\end{figure}

We have reduced the dimensionality of latent features as 3 using principal component analysis (PCA) to visualize our results. As visualization results for experimental results using in Figure \ref{fig3}, the weight matrix of a neural network composed of the column vectors which take linear independence, shows better discriminative power in their distribution of latent features than the neural network did not force the linear independence during network training.

\begin{figure*}
    \centering
    \begin{subfigure}{0.33\textwidth}
        \includegraphics[width=\textwidth]{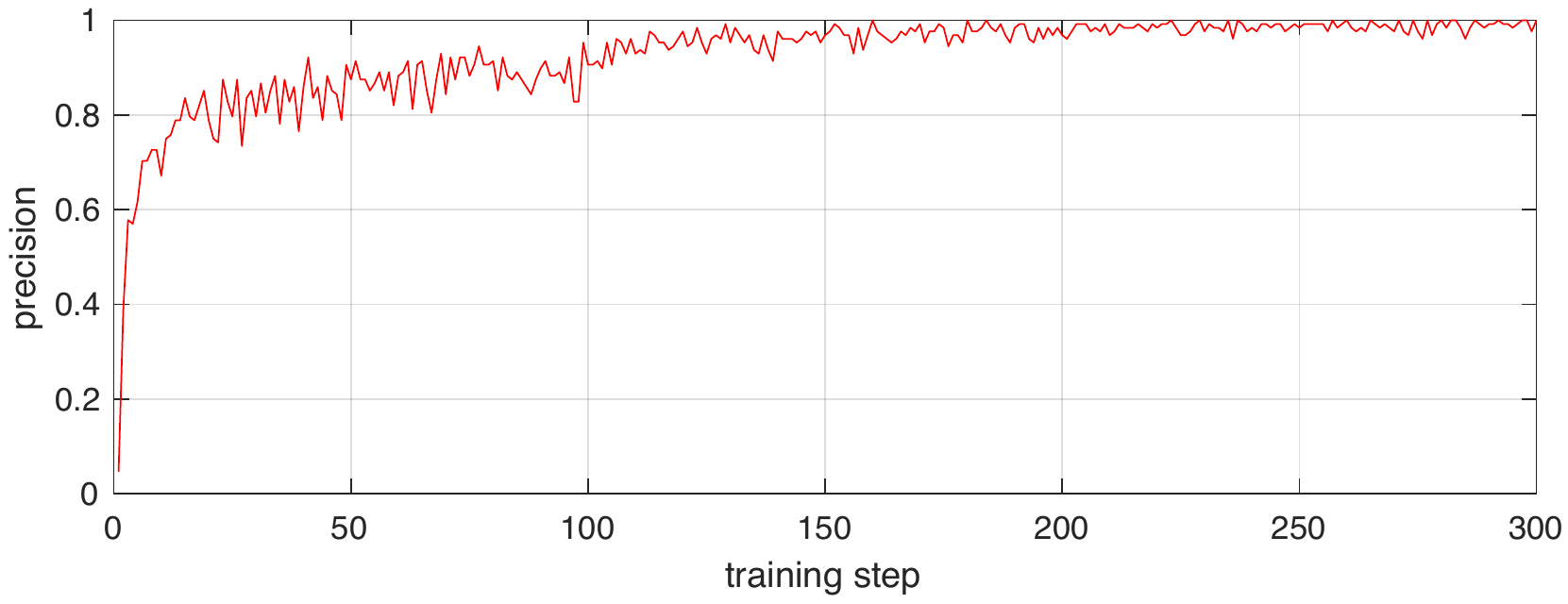}
\label{fig:precision}  
    \vspace{-2ex}
\caption{}
\label{fig:123:a}
    \end{subfigure}
    \hfill
    \begin{subfigure}{0.33\textwidth}
        \includegraphics[width=\textwidth]{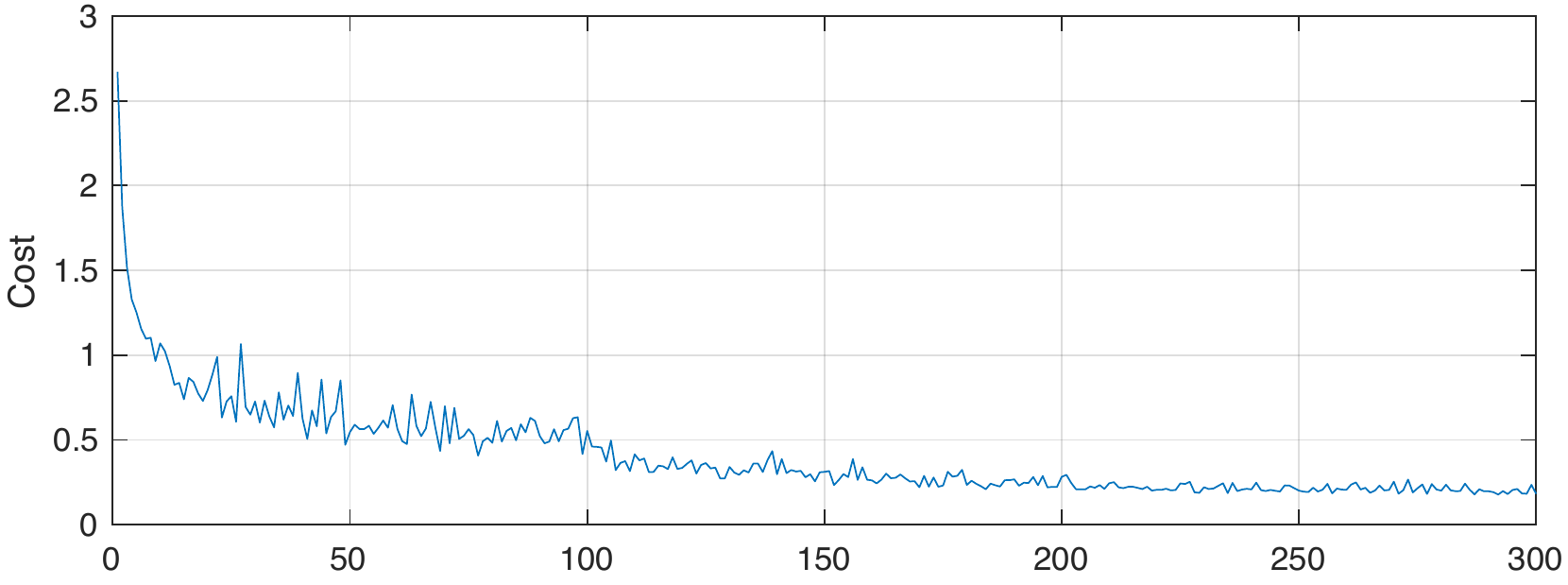}
    \label{fig:cost}  
    \vspace{-2ex}
\caption{}
 \label{fig:123:b}
    \end{subfigure}
    \hfill
    \begin{subfigure}{0.33\textwidth}
        \includegraphics[width=\textwidth]{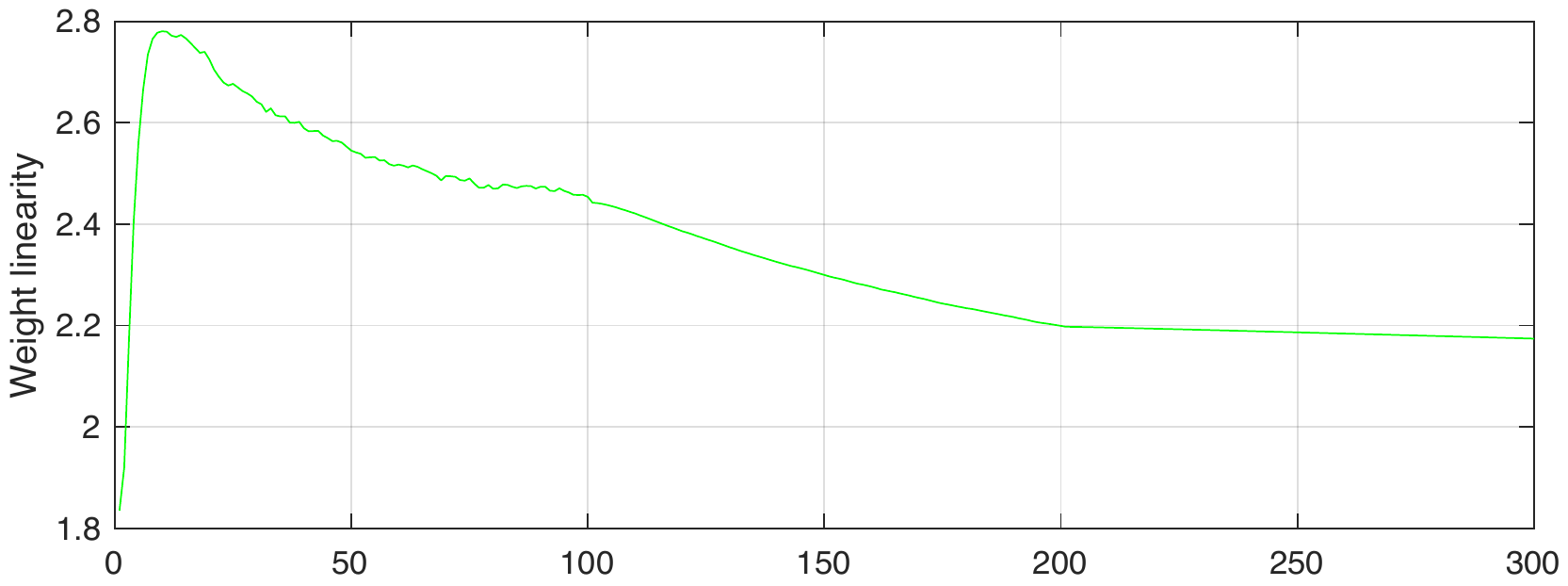}
         \label{fig:lin}  
    \vspace{-2ex}
\caption{}
\label{fig:123:c}
    \end{subfigure}
    \hfill
    \caption{(a), (b) and (c) contains the classification accuracies, costs, and the kernel linearity ($\epsilon(W)$) on each training step respectively. $X$-axis of each graph denote the training step. The baseline model is ResNet-32.}
\label{fig:v5}   
\vspace{-2ex}
\end{figure*}

\section{Weight Separability Evaluation}

\subsection{Intuition}
As the illustration in Fig \ref{fig2} and the experimental results in Fig \ref{fig3}, the linearity of the column vectors in a weight matrix can influence recognition performances. We try to evaluate the weight separability using the orthogonality of a matrix. The property of orthogonal matrix is as follows: $QQ^{T} = Q^{T}Q = I$, where $Q$ is a square matrix, and $I$ is a corresponding identity matrix of $Q$. However, the dimensionality of the commonly used weight matrix $W$ is not a square matrix, and also we can not guarantee that the weight matrix $W$ is invertible in practical situations. Therefore, in this work, we employ the concept of a semi-orthogonal matrix. A non-square matrix $A$ is semi-orthogonal if either $AA^{T} = I$ or $A^{T}A = I$, and it implies that $A$ take isometry property. With this notation, the linearity of a weight matrix $W\in R^{m\times{}n}$ is simply evaluated by calculating an error $E$ defined as follows:
\begin{equation}
\begin{aligned}
E(W,I) = W^{T}W - I_{n}, E(W,I)\in R^{n \times n},
\end{aligned}
\end{equation}
where $W$ is a weight matrix and $I_{n}$ is the corresponded identity matrix of $n\times{}n$ dimension. The result of this subtraction operation is a matrix. When $E(W, I)$ are closer to a zero matrix, $W$ can take stronger linearity. However, matrix form is inappropriate to consider as a quantitative value to estimate the linearity. Moreover, in practice, Above equation does not show the complete equivalence as mathematical semi-orthogonal. The cause of this inequivalence is a matrix structure of a neural network. The matrix notation for a final fully connected network is represented as follow:
\begin{equation}
\begin{aligned}
\boldsymbol{\alpha} \cdot [w_{1},w_{2},w_{3},...,w_{n}]= o,
\end{aligned}
\end{equation}
where $\alpha\in R^{1\times m}$ is the latent feature outputed from a previous layer which consisting of $m$ of elements,  $w_{i}\in R^{m\times1}$ is $i^{th}$ column vector in weight matrix $W$ of the final layer, and $o\in R^{1\times n}$ is the output of network. $n$ is the number of classes. In above notation, each output $o_{i}$, where $i=1,2,3,...,n$, is calculated as follows:
\begin{equation}
\begin{aligned}
o_{i} = \alpha\cdot w_{i}=\sum^{m}_{j=1}\alpha^{j}w_{ij},
\end{aligned}
\end{equation}
where $w_{ij}$ is $j^{th}$ element of the $i^{th}$ column vector $w_{i}$. In the above notations, the column vectors in weight matrix play a rule as a kernel to assign specific class by computing vector similarity between the given feature $\alpha$ and each column vector $w^{f}_{i}$. In this work, we consider the separability of weight kernel so that we only consider the linear independent of column vectors of weight matrix $W$. However, this principle can be used for the network in which their row vector is used for the decision kernel.

\begin{figure*}
	\vspace{-0.2cm}
	\centering
\includegraphics[width=\textwidth]{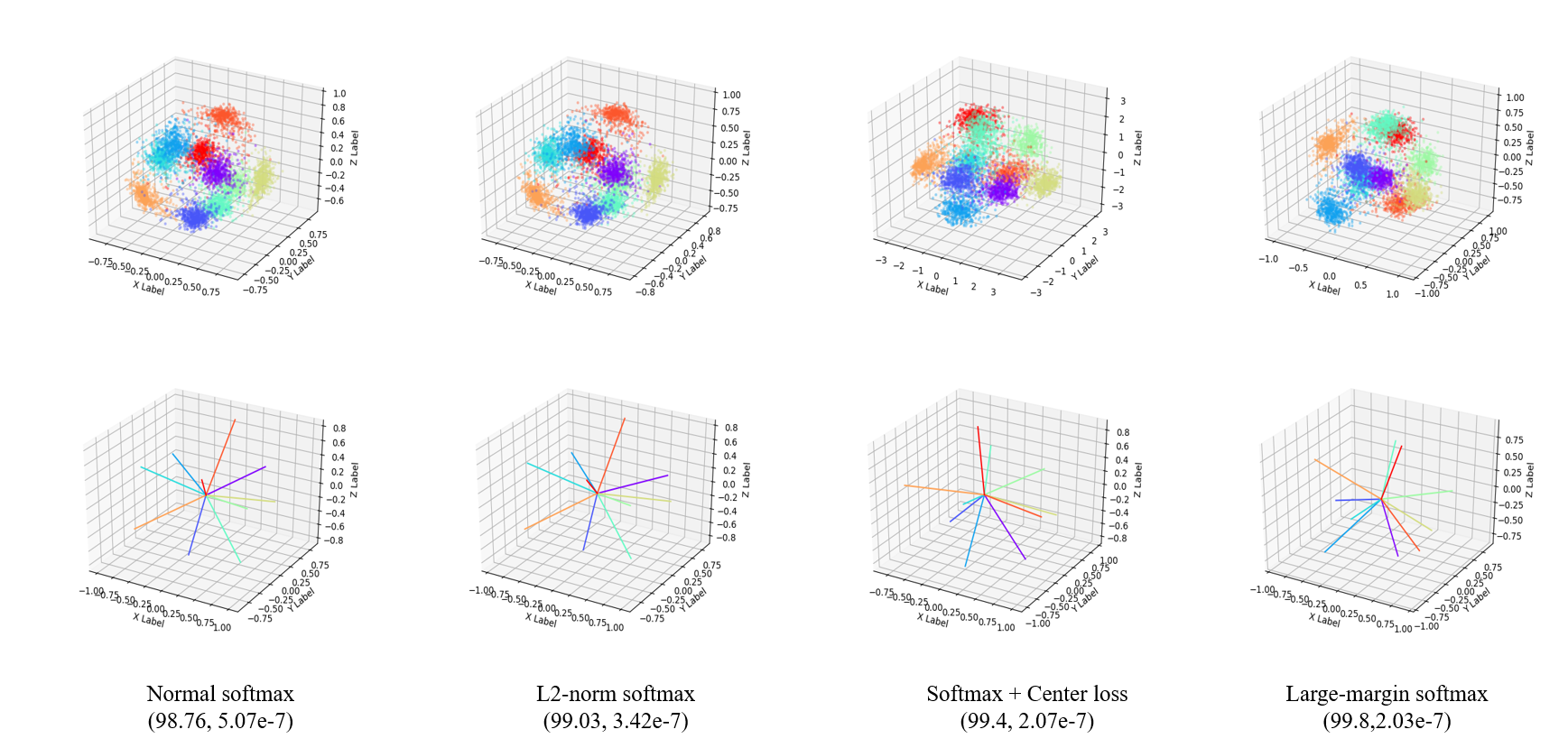}
	\caption{The visualization results for latent features and learned weight vectors of models trained by various loss functions. The graphs in the first row show the distribution of latent features in 3D space. The graphs in the second row represent the direction and magnitude of weight vectors. We employ the LeNet structure in this visualization, and we reduce the dimensionality of latent features as 3 using PCA. The figures under the name of loss functions show the classification accuracies and the results of weight separability evaluation using our metric. The noticeable thing is that recognition performance and the results of weight separability are proportion, even though the visualization results are difficult to correspond to the recognition performance.}
	\label{fig4}
	\vspace{-0.2cm}
\end{figure*}

\subsection{Metric Definition and Mathematics}
Since a matrix format in Eq 3. is not suitable to evaluate the weight separability quantitatively, we employ Frobenius Distance which can be converting the matrix form to real-number. We define the quantitative metric based on Frobenius Distance to evaluate the linearity of column vectors in a weight matrix. The metric $\epsilon(W)$ for separability of a weight matrix $W \in R^{m \times n}, m>n$ is defined by
\begin{equation}
\begin{aligned}
\epsilon(W) = \frac{1}{n}\parallel{}W^{T}W - I_{n}\parallel{}_{F}^{2}  
\end{aligned}
\end{equation}
$n$ is the number of column vectors in the weight matrix, and $I_{n}$ is an identity matrix with $n\times n$ dimension. The proposed metric computes the weight separability using Frobenius distance and regularizes it by dividing with the number of classes. The reason for the regularization with the number of classes is to provide the generalized evaluation metric invariant to the number of classes, and prevent the fluctuating evaluation values according to the problem domain. In equation 6, $W^{T}W-I$ is represented as follows:
\begin{equation}
\begin{aligned}
\begin{bmatrix}
    w_{11} & \dots  & w_{1n} \\
    w_{21} & \dots  & w_{2n} \\
    \vdots &  \ddots & \vdots \\
    w_{m1} &  \dots  & w_{mn}
\end{bmatrix}^{T}
\begin{bmatrix}
    w_{11} & \dots  & w_{1n} \\
    w_{21} & \dots  & w_{2n} \\
    \vdots &  \ddots & \vdots \\
    w_{m1} &  \dots  & w_{mn}
\end{bmatrix}
-
\begin{bmatrix}
    1 & \dots  & 0 \\
    \vdots &  \ddots & \vdots \\
   0 &  \dots  & 1
\end{bmatrix}
\\=
\begin{bmatrix}
    \sum_{i=1}^{m}w_{i1}^{2}-1  & \dots  &  \sum_{i=1}^{m}w_{i1}w_{in} \\
    \sum_{i=1}^{m}w_{i1}w_{i2} & \dots  & \sum_{i=1}^{m}w_{i2}w_{in} \\
    \vdots  & \ddots & \vdots \\
    \sum_{i=1}^{m}w_{i1}w_{in} & \dots  & \sum_{i=1}^{m}w_{in}^{2}-1
\end{bmatrix} \in R^{n\times n},
\end{aligned}
\end{equation}
where $w_{ij}$ is $i^{th}$ row and $j^{th}$ column element in a weight matrix. By the properties of transpose: 1) $(A^{T})^{T}=A$ and 2) $(A-B)^{T}=A^{T}-B^{T}$, the result of $W^{T}1^{T})^{T}-I=W^{T}W-I$. By this property, the metric in Eq. (6) can be represented as follows:
\begin{equation}
\begin{aligned}
e(W) =\frac{1}{n}Tr((W^{T}W - I_{n})^{T}(W^{T}W - I_{n}))\\=\frac{1}{n}Tr((W^{T}W - I_{n})^{2}).
\label{eq:toto}
\end{aligned}
\end{equation}
where $Tr(\cdot)$ is the trace operation of an square matrix defined by the sum of the elements on the main diagonal of the square matrix. Intuitively, when the value of $e(W)$ is converged to zero, the column vectors of weight matrix would be linearly independent and separability of the column vector can take maximum. We omit the bias in the fully connected layer because it just complicates our analysis based on visualization and nearly does not influence the recognition accuracies \cite{liu2016large}. Figure \ref{fig:v5} shows the trend of the classification precision, cost function, and the kernel linearity evaluated by Eq. \ref{eq:toto}, based on ResNet-32 and Cifar-10 dataset. As shown in figure \ref{fig:v5}, the kernel linearity is gradually decreased and the classification precision increasing during the training.

Additionally, we conducted simple experiments using MNIST dataset to verify our metric. We trained the LeNet using various loss functions including $l{2}$-norm softmax \cite{taigman2014deepface}, center loss\cite{wen2016discriminative}, and large-margin softmax \cite{liu2016large}, and carried out the cross check for accuracy and weight separability about each model. Figure \ref{fig4} illustrates the visualization results of the experiments. As the results in Fig \ref{fig4}, the experimental results show that the more accurate recognition performance can take the larger weight separability evaluated as our metric. One of the interesting observations is that the evaluation results for weight separability using our metric can be reflected the recognition performance, even if it is difficult to figure out the superiority of recognition performance using visualization results.

\begin{figure*}[ht]
	\vspace{-0.2cm}
	\begin{center}
        \includegraphics[width=\textwidth]{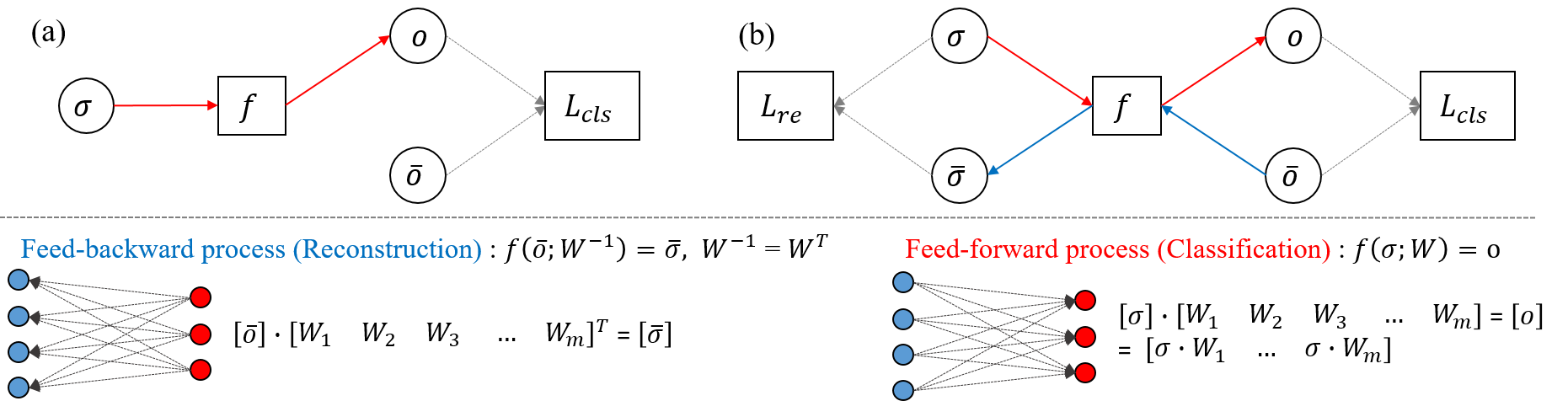}
		\caption{(a) The normally trained models contain a simple mapping pipeline for classification $f$, and associated classification loss $L_{cls}$. (b) The models applied the proposed reconstruction loss contains two mapping pipelines: the classification $f$ and reconstruction $f^{-1}$, and associated losses $L_{cls}$, $L_{re}$ for each. $o$ and $\hat{o}$ are the network output and corresponding annotation. $\alpha$ and $\hat{\alpha}$ is the latent feature and reconstructed latent feature from the given annotation $\hat{o}$ respectively. The red and blue arrows in first row indicate the classification and reconstruction pipelines respectively. The red and blue dots represent the activation units of output and previous layers.}
		\label{fig5}
	\end{center}
	\vspace{-0.2cm}
\end{figure*}

\section{Feed-backward Reconstruction}
\subsection{Motivation}
Consider the commonly used optimization methods such as softmax-cross entropy, and we have a latent feature $\alpha$ and corresponding annotation label $o$. If the latent feature $\alpha$ have to classified to $i^{th}$ class, then the methods are concentrate on to encourage $w_{i}\cdot\alpha > w_{j}\cdot\alpha, j=1,2,3,4, ..., n$ and $j\neq i$, where $n$ is the number of classes, and $w_{i}$ is $i^{th}$ column vector in a weight matrix $W$. In this work, we want to improve not only intra-class compactness and inter-class separability, but also boost the separability between the column vectors in weight vector. Current loss functions such as softmax-cross entropy, $l_{2}$-distance loss, cosine angular loss, and large-margin softmax, do not consider the weight separability explicitly. So the new method is required to directly improve the weight separability.

\subsection{Feed-backward Reconstruction Loss}
Following the notation for the weight separability evaluation in Section 3, the weight separability would be maximum when $W^{T}W-I$ takes a zero matrix. In this case, basically, we assume that $W^{T}=W^{-1}$. However, using the proposed evaluation metric as an objective function is unsuitable to train a model because of a problem for computing gradient as long as we use the back-propagation algorithm \cite{hecht1992theory} to update network parameters. The evaluation metric is composed of the weight matrix of the final layer only, therefore the gradient of the proposed metric for weight separability $\frac{\partial{}\frac{1}{n}Tr((W^{T}W - I_{n})^{2})}{\partial{}w_{ij}}$, will vanishes when the gradient for other layers are calculated. Consequently, it is necessary to develop an objective function which is suitable for applying the trainig procedure of networks. 

To address this issue, we propose the feed-backward reconstruction loss that can improve the weight separability directly. The feed-backward reconstruction loss is defined as
\begin{equation}
\begin{aligned}
L_{re}(\hat{o};\alpha,W) = \sum_{i=1}P(\alpha_{i})\text{log}(\frac{P(\alpha_{i})}{Q(\hat{o}w_{i}^{T})})
\end{aligned}
\end{equation}
where $\alpha$, $w_{i}^{T}$, and $\hat{o}$ ,are a latent feature, the $i^{th}$ transposed column vector of the weight matrix $W$ , and corresponding label about the latent feature. $P$ and $Q$ are the distributions for the latent features and reconstruction results. The proposed loss functions mathematically equivalent to the KullbackLeibler divergence, and literally this loss function defines the difference between the distributions of latent feature and reconstruction results. Intuitively, if the proposed loss $L_{re}$ is converged to zero, then it means $P(\alpha_{i})$ is equivalent to $Q(\hat{o}w_{i}^{T})$, and it is represented as, $P(\alpha)log(\frac{P(\alpha)}{Q(\hat{o}W^{T})})=0$. In this situation, $W^{T}$ can be regarded as $W^{-1}$ and it also can be regarded as a solution to maximizing the weight separability. The reconstruction loss functions using $l_{1}$-norm or $l_{2}$-norm force to minimize the Euclidean distance even their angular difference is tiny. These approaches can not be used with various activation functions since there is a probability that the Euclidean distance can be changed by an activation function. Therefore, so we instead require parameter transformations invariant method based on computing a difference of probabilistic distribution. 

When we apply the proposed loss to train a model, the proposed loss is added to ordinary loss functions $L_{cls}$ such as softmax cross entropy, center loss \cite{wen2016discriminative}, and large-margin softmax loss \cite{liu2016large}. Therefore, the total loss function is defined as follows, 
\begin{equation}
\begin{aligned}
L_{total}(\hat{o},o;\alpha,\theta) = L_{cls}(\hat{o},o;\theta)+\lambda{} L_{re}(\hat{o};\alpha,W),
\end{aligned}
\end{equation}
where $o$ and $\hat{o}$  are the output of models and corresponding labels. $\alpha$ is the output of previous layer that connected to the network for recognition tasks, and $W$ is the weight of a final layer. $\theta$ is a set of network parameters including $W$. $\lambda$ is hyper-parameter to decide the weight of the proposed reconstruction loss in training task. In our experiments, the value of $\lambda$ is set to 0.001, and this value is determined by the value with the best performance from several experiments.

\begin{table*}[]
\centering
\resizebox{1.0\textwidth}{!}{%
\begin{tabular}{l|cc|ccc|ccc}
\hline
Method & Depth & Params & C10 & C10+ &$e(W)^{avg}_{C10}$  & C100 & C100+ &$e(W)^{avg}_{C100}$  \\ \hline\hline
Network in Network \cite{lin2013network} & 12 & 11.4M & 13.76 & 11.2 & 8.56e-08 & 35.68 & 33.04 & 6.85e-08 \\
Network in Network$+L_{re}$  & 12 & 11.4M & 10.03({\color{red}-3.73}) & 9.64({\color{red}-1.56}) & 6.29e-08({\color{red}-2.27e-08})  & 31.22({\color{red}-4.46}) & 31.07({\color{red}-1.97}) & 6.81e-08({\color{red}-0.04e-08})  \\\hline
VGG-16 \cite{simonyan2014very} & 16 & 13.4M & 10.48 & 10.26 & 7.32e-08 & 37.48 & 31.27 & 6.51e-08 \\
VGG-16$+L_{re}$ & 16 & 13.4M & 9.17({\color{red}-1.31}) & 7.94({\color{red}-2.32}) & 6.07e-08({\color{red}-1.25e-08})  & 31.55({\color{red}-5.93}) & 29.96({\color{red}-1.31}) & 6.43e-08({\color{red}-0.08e-08}) \\\hline
Highway Network \cite{srivastava2015training} & 12 & 11.8M & 12.98 & 9.8 & 7.32e-08 & 39.51 & 33.07 & 5.12e-08 \\ 
Highway Network$+L_{re}$ & 12 & 11.8M & 9.13({\color{red}-3.85}) & 7.72({\color{red}-2.08}) & 6.73e-08 ({\color{red}-0.59e-08})  & 35.64({\color{red}-3.07}) & 32.01({\color{red}-1.06}) & 4.15e-08({\color{red}-0.97e-08}) \\ \hline
ResNet-32 \cite{he2016deep} & 36 & 1.7M & 8.64 & 8.09 & 6.55e-08 & 32.18 & 31.37 & 4.62e-08 \\
ResNet-32$+L_{re}$  & 36 & 1.7M & 6.01({\color{red}-2.63}) & 5.94({\color{red}-2.15}) & 3.57e-08({\color{red}-2.98e-08}) & 30.65({\color{red}-1.03}) & 29.48({\color{red}-1.89}) & 2.96e-08({\color{red}-1.66e-08}) \\ \hline
DenseNet-40 $(k=12)$ \cite{huang2017densely} & 40 & 1.0M & 9.42 & 6.17& 3.45e-08 & 29.3 & 24.58 & 1.54e-08 \\
DenseNet-40$+L_{re}$ $(k=12)$  & 40 & 1.0M & \textbf{5.91}({\color{red}-3.51}) & \textbf{5.62}({\color{red}-0.55}) & 2.16e-08({\color{red}-1.29e-08}) & \textbf{29.01}({\color{red}-0.29}) & \textbf{20.75}({\color{red}-3.83}) & 1.42e-08({\color{red}-0.12e-08}) \\ \hline
\end{tabular}%
}
\caption{Error rates (\%) on CIFAR-10 and CIFAR-100 datasets. $+L_{re}$ denotes the model is trained with the proposed reconstruction error. $+$ indicates that simple data augmentation is used. $e(W)_{avg}$ is the average result of weight separability evaluation between normally trained results and the results with the simple data augmentation corresponding to C10 and C100 dataset. $k$ is the growth rate in DenseNet. + indicates that the data augmentation based on simple image transformation is used. The marked value as red colour is a change of performance after applying the proposed reconstruction loss. The bolded value is the best performance in our experiments.}
\label{my-label}
\vspace{-1ex}
\end{table*}

\subsection{Interpretation}
The model applied the feed-backward reconstruction loss contains two mapping process: 1) Determination process $f:\alpha \xrightarrow{}o$ and 2) Reconstruction process  $f^{-1}:\hat{o}\xrightarrow{} \hat{\alpha}$, and both processes share weight parameter $W$. The determination process $f$ encourages $W$ to translate $\alpha$ into an encoded output $o$, and the reconstruction process $f^{-1}$ force $W^{T}$ to recover $\hat{\alpha}$ from given label $\hat{o}$. Figure \ref{fig5} shows the comparison between a normal model and the model applying the feed-backward reconstruction process in a classification task. In optimization via these two processes, each process affects each other in achieving their objectives.

The objective of the determination process is to maximize the accuracy for visual recognition tasks by minimizing geometric or probabilistic difference between the output of a model $\alpha{}W=o$ and the given annotations $\hat{o}$. The reconstruction process aims to minimize the difference of distributions between the latent feature $P(\alpha)$ and the reconstruction results $Q(\hat{o}W^{T})$. The reconstruction process can be optimized when the determination process takes highly accurate performance, and it is able to provide more accurate recognition performance when the weight separability become more advanced. Above cooperation between two processes is similar to the cycle consistency losses \cite{zhu2017unpaired}. Consequently, above processes not only can boost the weight separability but also can improve the cyclic consistency via dual minimization schemes for classification task and latent feature reconstruction.

\begin{table*}[t]
\centering
\footnotesize
\resizebox{0.9\textwidth}{!}{%
\begin{tabular}{c|c|cc|cc}
\hline
Method  & Data & LFW & $e(W)_{LFW}$ & YTF & $e(W)_{YTF}$ \\
\hline\hline
DeepFace \cite{taigman2014deepface}& WebFace & 3.65 & 11.67e-08 & 13.77& 14.01e-08 \\
DeepFace$+L_{re}$ & WebFace & 2.99({\color{red}-0.66})& 8.43e-08({\color{red}-3.24e-08}) & 10.24({\color{red}-3.53})& 11.54e-08({\color{red}-2.47e-08}) \\\hline
FaceNet \cite{schroff2015facenet}& WebFace& 2.82 & 9.78e-08 & 6.21& 9.76e-08 \\
FaceNet$+L_{re}$  & WebFace & 2.60({\color{red}-0.22}) & 8.96e-08({\color{red}-0.82e-08}) & 7.87({\color{red}-0.34}) & 9.13e-08({\color{red}-0.63e-08}) \\\hline
DeepID \cite{sun2015deeply} & WebFace & 3.08 & 10.03e-08 & 7.35 & 10.83e-08\\
DeepID$+L_{re}$   & WebFace & 1.66({\color{red}-1.42}) & 8.01e-08({\color{red}-2.02e-08})& \bf{4.53}({\color{red}-2.82})& 8.76e-08({\color{red}-2.07e-08})\\\hline
DDRL \cite{yu2018deep} & WebFace & 0.99 & 6.81e-08 &5.98 & 10.91e-08\\
DDRL$+L_{re}$ & WebFace & \bf{0.87}({\color{red}-0.12})& 6.30e-08({\color{red}-0.51e-08}) & 7.15({\color{blue}+1.17})& 7.38e-08({\color{red}-3.54e-08}) \\\hline
L-Softmax \cite{liu2016large}& WebFace & 1.48& 7.74e-08 & 6.21 & 9.65e-08\\
L-Softmax$+L_{re}$ & WebFace & 0.94({\color{red}-0.54})& 6.84e-08({\color{red}-0.90e-08}) & 5.57({\color{red}-0.64})& 8.68e-08({\color{red}-0.97e-08}) \\\hline
Softmax+Center Loss \cite{wen2016discriminative} & WebFace & 1.22& 10.24e-08 & 6.08& 13.42e-08\\
Softmax+Center Loss$+L_{re}$ & WebFace & 1.47({\color{blue}+0.25})& 9.53e-08({\color{red}-0.71e-08}) & 6.03({\color{red}-0.05})& 11.97e-08({\color{red}-1.45e-08})\\
\hline
\end{tabular}
}
\caption{\footnotesize Error rate (\%) and the results of weight separability evaluation using our metric ($e(W)$) on LFW and YTF datasets. $+L_{re}$ denotes the model is trained with the proposed reconstruction error. $+$ and $-$ represent that the increase or decrease on recognition error rate after applying the proposed reconstruction loss. $e(W)_{LFW}$ and $e(W)_{YTF}$ indicate the evaluation result  of the proposed metric for weight separability for each dataset.  For a fair comparison, we implemented all models and loss functions directly and trained only using CASIA-Webface dataset. The bolded values represent the lowest error rate on LFW and YTF datasets.}
\vspace{-2ex}
\end{table*}

\section{Experimental results}
\subsection{Image Classification}
We conducted experiments for image classification on the CIFAR-10 and CIFAR-100 datasets \cite{krizhevsky2009learning}. The CIFAR-10 dataset is composed of 50,000 training images and 10,000 test images in 10 classes. CIFAR-100 dataset consists of 100 classes, and each class contain 500 training images and 100 testing images. Our work is concentrated to demonstrate the efficiency of the feed-backward reconstruction loss, and not on encourage state-of-the-art performance. Therefore, our experiment conducted based on the several baseline models intentionally and focused on the comparison between normally trained model and trained model using the feed-backward reconstruction loss .

The baseline models used in the experiment for image classification, are as follows: Network in Network \cite{lin2013network}, VGG-16 \cite{simonyan2014very}, Highway Network \cite{srivastava2015training}, Residual Network (ResNet) \cite{he2016deep}, and Densely Connected Convolutional Neural Network (DenseNet) \cite{huang2017densely}. To improve an experimental efficiency, we use the most shallow structure on ResNet and DenseNet, and the ResNet-32 and Densenet-40 structures are selected for our experiments.  All networks are trained using stochastic gradient descent (SGD) \cite{bottou2010large}. We trained all networks using 128 batch size for 300 epochs. During training networks, we employed learning rate decay of 0.0001 and momentum of 0.9. The learning rate is initially set to 0.1, and divided by 10 in 100, 200, and 250 epochs.

The experimental results on CIFAR-10 and CIFAR-100 dataset are shown in Table 1. The densely connected convolutional network applying simple data augmenation and the proposed reconstruction loss achieved an error rate of 5.62\% on CIFAR-10 dataset and 20.75\% on CIAR-100 dataset. These figures are the best results in our experiment for image classification. The evaluation results of weight separability for these experiments are 2.16e-08 and 1.42e-08 respectively. The experimental results show that the trained model considering the feed-backward reconstruction loss outperformed the normally trained models. The most noticeable things in our experiment are that the models trained reflecting our loss achieve better performance whether the performance differences are small or large collectively.

\subsection{Face Recognition}
We have conducted additional experiments for face recognition to demonstrate the efficiency of the proposed method for improving weight separability. This experiment is conducted under the \textit{unrestricted with labelled outside data} protocol, so that all models were trained only using CASIA-Webface dataset and tested using Labeled Faces in the Wild (LFW) dataset \cite{huang2007labeled} and the Youtube Faces (YTF) \cite{wolf2011face} dataset. CASIA-Webface dataset consists of 494,414 of face images labelled as 10,575 different identities, and the dataset also contains horizontally flipped images for data augmentation. The performance evaluation is carried out on 6000 of face pairs from LFW dataset, and 5000 of video pairs from YTF dataset. 

The network model list used in this experiments as follows: DeepFace \cite{taigman2014deepface}, Facenet \cite{schroff2015facenet}, DeepID2+ \cite{sun2015deeply}, DDRL \cite{yu2018deep}, and the other methods proposed by Wen \etal \cite{wen2016discriminative}, and Liu \etal \cite{liu2016large}. These methods are initially trained via classification setting and conduct the evaluation using a verification scheme. We added the feed-backward reconstruction loss in calculating the total loss when the models are trained. Table 2. shows the comparison results of the normally trained models and the models applying the proposed loss.

The face recognition results usually show that the trained models applying the proposed loss achieved better performance than the normally trained models. The highest recognition accuracies in LFW and YTF datasets are achieved by the DDRL and DeepID frameworks trained with the proposed reconstruction loss. These models achieve 0.87\% and 4.53\% error rates on LFW and YTF datasets respectively. The evaluation results of weight separability for these experiments are 6.30e-08 and 8.76e-08.  However, in experiments using the DDRL and the center loss, the proposed method degraded the recognition accuracies. In the experiment using YTF dataset and DDRL, the 3.54e-08 of weight separability was reduced, but the DDRL applying the proposed reconstruction loss have achieved 7.15$\%$, and this figure is lower than then 5.98$\%$ of the original model. Additionally, the experiment using the center loss, the trained model with the proposed reconstruction loss achieved lower accuracies than the original model.

The overall experimental results on face recognition tasks show similar trend on the experimental results of image classification. Even though the experimental results in our experiment are slightly lower then the listed accuracies in their studies, these figures are comparable to the reported performance in the studies \cite{schroff2015facenet,sun2015deeply,liu2017sphereface} and almost similar to the state-of-the-art methods only trained by CASIA-Webface dataset. 

\begin{figure}[h]
	\centering
	\includegraphics[width=\columnwidth]{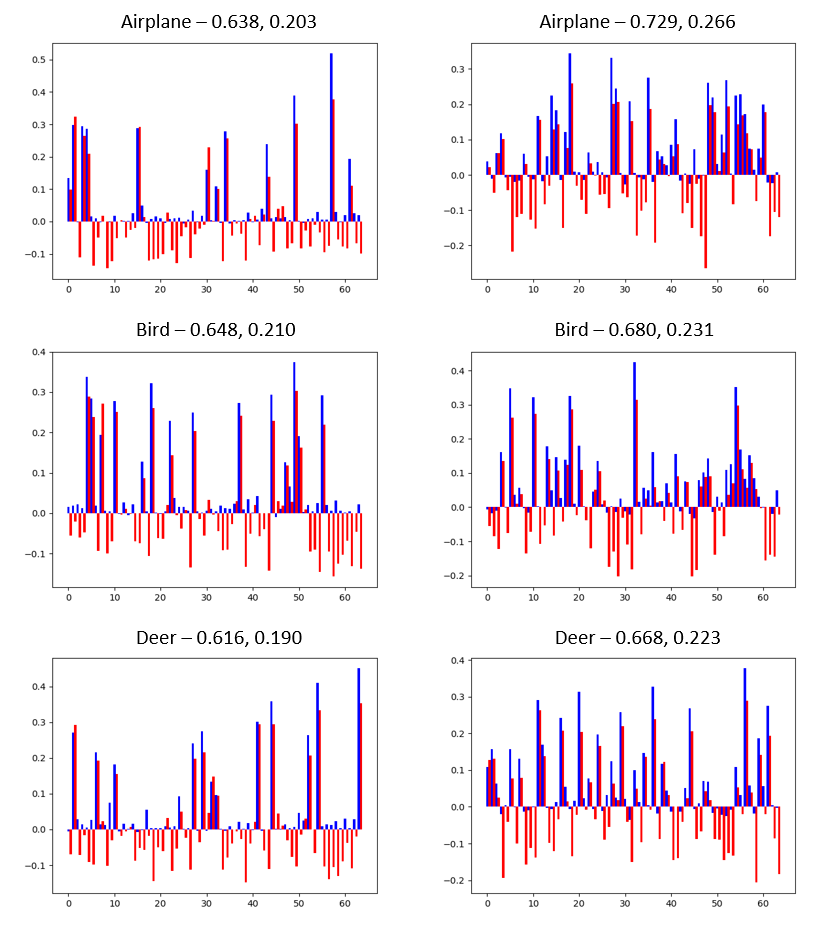}
	\caption{Pattern comparison of neuron activation and the corresponding weight vector on 'Airplane', 'Bird', and 'Deer' classes in CIFAR-10 dataset. X-axis shows the an index of each neuron, and Y-axis represents an activation output. The graphs in right-side are the pattern comparison for normally trained ResNet, and the graphs in left-side are the comparison on the ResNet applying the proposed reconstruction loss. The blue bar indicates the expectation of neuron activation, and the red bar represents the corresponding weight vector. The values beside of class name are vector similarities based on Euclidean distance and cosine similarities between the expectation value of neuron activation and the corresponding weight vector.}
	\label{fig6}
	\vspace{-2ex}
\end{figure}

\subsection{Analysis}
The experimental results show clear advantages over current deep neural network models and a lot of compared baselines. Our interpretation of these performance improvements is as follows. In first, as we mentioned in Section 2 and Section 3, the weight separability can influence recognition performance in a model based on the neural network. We tried to improve the weight separability via feed-backward reconstruction loss which can encourage the linear independence between the column vectors in a weight matrix. In the learning procedure, the proposed reconstruction loss plays an important role to improve the weight separability explicitly. The error rates and weight separability evaluation results in Table 1, show that the classification performance is probably proportional to the weight separability evaluation results. Not only image classification results, but also experimental results for face recognition shows similar circumstance. 

In Second, the feed-backward reconstruction can improve the not only weight separability but also intra-class compactness. Figure \ref{fig6} represents the comparison of neuron activation pattern and the values of a corresponding column vector in a weight matrix in our classification experiment using ResNet. The figures on the top of a bar graph indicate that the Euclidean distance and cosine similarity between the neural activation and the corresponding column vector in a weight matrix.

These figures are regarded as that the similarities between neuron activation and the corresponding vectors. A common point of these figures is the figures applying the proposed reconstruction loss, are smaller than the normal ones. In figure 6, the Euclidean distance and cosine similarity of the model applying our reconstruction loss, about 'Deer' class are 0.616 and 0.190. On the contrary, the corresponding Euclidean distance and cosine similarity of the normal model are 0.666 and 0.223, and these figures are bigger than the model applying the proposed reconstruction loss. In addition to the experimental results for 'Deer' class, Other experimental results for 'Airplane' and 'Bird' classes shows the same phenomenon. These results show that the proposed reconstruction loss can help to learn more discriminative representation.

\section{Conclusion}
In this paper, we presented the metric for weight separability evaluation and proposed the feed-backward reconstruction loss to directly improve the weight separability which can be used for various visual recognition tasks. The evaluation metric for weight separability can represent linear independence property of column vectors in a weight matrix. With feed-backward reconstruction loss, the separability of column vectors in weight matrix was improved. The experimental results present that the proposed feed-backward process and the loss function significantly contribute performance improvement in recognition tasks.   

\small
\bibliographystyle{aaai}
\bibliography{refs}

\end{document}